\documentclass[preprint,12pt]{elsarticle}
\usepackage{amssymb}
\usepackage{amsmath}
\usepackage{multirow}
\usepackage{array}
\usepackage{booktabs}
\usepackage{adjustbox}
\usepackage{url}

\journal{Biomedical Signal Processing and Control}

\begin{document}

\begin{frontmatter}

\title{MMSF: Multitask and Multimodal Supervised Framework for WSI Classification and Survival Analysis}

\author[1,2]{Chengying She}
\author[3]{Chengwei Chen}
\author[1,2]{Xinran Zhang}
\author[1,2]{Ben Wang}
\author[2,4]{Lizhuang Liu\corref{cor1}}
\author[3]{Chengwei Shao\corref{cor2}}
\author[3]{Yun Bian\corref{cor2}}
\cortext[cor1]{Corresponding author. E-mail: liulz@sari.ac.cn}
\cortext[cor2]{Corresponding author. E-mail: chengweishaoch@163.com, bianyun2012@foxmail.com}

\affiliation[1]{organization={University of Chinese Academy of Sciences},
            city={Beijing},
            country={China}}
\affiliation[2]{organization={Shanghai Advanced Research Institute, Chinese Academy of Sciences},
            city={Shanghai},
            country={China}}
\affiliation[3]{organization={Department of Radiology, Changhai Hospital},
            city={Shanghai},
            country={China}}
\affiliation[4]{Lead contact}

\begin{abstract}
    Multimodal evidence is critical in computational pathology: gigapixel whole slide images capture tumor morphology, while patient-level clinical descriptors preserve complementary context for prognosis. Integrating such heterogeneous signals remains challenging because feature spaces exhibit distinct statistics and scales. We introduce MMSF, a multitask and multimodal supervised framework built on a linear-complexity MIL backbone that explicitly decomposes and fuses cross-modal information. MMSF comprises a graph feature extraction module embedding tissue topology at the patch level, a clinical data embedding module standardizing patient attributes, a feature fusion module aligning modality-shared and modality-specific representations, and a Mamba-based MIL encoder with multitask prediction heads. Experiments on CAMELYON16 and TCGA-NSCLC demonstrate 2.1--6.6\% accuracy and 2.2--6.9\% AUC improvements over competitive baselines, while evaluations on five TCGA survival cohorts yield 7.1--9.8\% C-index improvements compared with unimodal methods and 5.6--7.1\% over multimodal alternatives.

    The project source code is publicly available at \url{https://github.com/chengyingshe/MMSF}.
\end{abstract}

\begin{graphicalabstract}
\centering
\includegraphics[width=\textwidth]{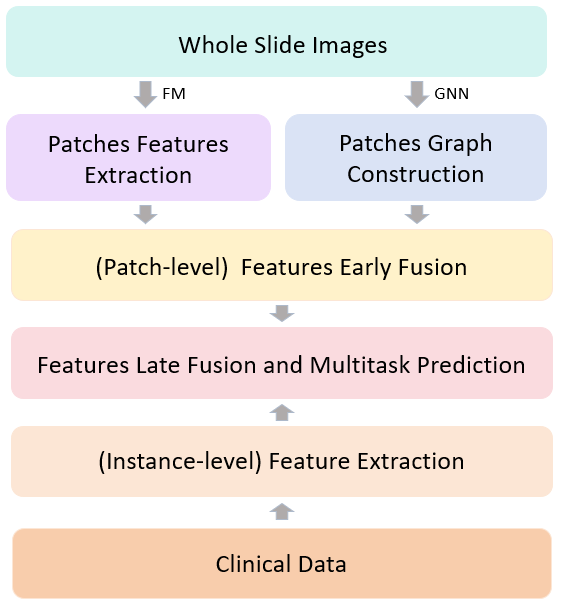}
\end{graphicalabstract}

\begin{highlights}
\item MMSF is the first linear-complexity multitask framework for WSI classification and survival analysis using state space models
\item MMSF proposes a novel clinical data embedding module to extract embedding features from clinical data for downstream tasks
\item MMSF introduces a hierarchical fusion mechanism combining patch-level graph features and instance-level clinical embedding features
\item MMSF achieves state-of-the-art performance on classification and survival analysis across multiple datasets, and it is evaluated on seven public datasets demonstrating generalizability and clinical applicability
\end{highlights}


\end{frontmatter}



\section{Introduction}
\label{sec:intro}

Cancer is a highly heterogeneous disease and remains a leading cause of mortality worldwide, where accurate diagnosis and prognosis are vital for effective treatment planning~\cite{meacham2013_tumour_heterogeneity_cancer,campanella2019_clinical_grade_computational_pathology}. Computational pathology has emerged as a transformative field, leveraging artificial intelligence to analyze whole slide images (WSIs) for downstream tasks such as classification and survival analysis~\cite{song2023_ai4computational_pathology}. WSIs contain rich morphological information but pose computational challenges due to their gigabyte-scale size and inherent complexity~\cite{cornish2012_wsi,lu2021_clam}.

Machine learning methods have achieved remarkable success in cancer diagnosis and prognosis, excelling in tumor classification, survival analysis, and treatment response assessment~\cite{song2023_ai4computational_pathology,ilse2018_abmil,wang2019_ml_for_survival_analysis}. The mainstream WSI analysis pipeline~\cite{cornish2012_wsi} typically involves: (1) segmenting tissue regions from background, (2) cropping into fixed-size small patches (typically $224 \times 224$ pixels), (3) extracting features with CNN-based models (such as ResNet~\cite{he2016_resnet}, DenseNet~\cite{huang2017_densenet}) or ViT-based models (such as DINO~\cite{zhang2022_dino}, DINOv2~\cite{oquab2024_dinov2}) that have been pre-trained on large-scale datasets like ImageNet~\cite{deng2009_imagenet} or specialized pathological image datasets, and (4) using multiple instance learning (MIL) algorithms for downstream tasks~\cite{lu2021_clam,chen2024_uni,ilse2018_abmil}. Despite progress, this pipeline faces several fundamental limitations.

The primary challenge lies in the inherently high-dimensional nature of WSIs, which are cropped into hundreds of thousands of patches that must be processed efficiently, leading to massive computational requirements that hinder clinical deployment. Transformer-based MIL models also suffer from quadratic complexity with respect to sequence length, further inflating computational costs~\cite{shao2021_transmil,jafarinia2024_snuffy}. Moreover, most approaches overlook spatial relationships between patches, which are key for capturing tissue architecture~\cite{jaume2021_histocartography}, and rely solely on pathological data while neglecting routinely collected clinical information. In contrast, methods integrating multi-omics data (e.g., genomics, transcriptomics) can improve accuracy but face barriers of cost and data availability~\cite{chen2021_mcat,chen2022_porpoise,zhou2023_cmta,fu2025_hsfsurv}.

To overcome these challenges, we propose MMSF, a Multitask and Multimodal Supervised-learning Framework for WSI classification and survival analysis. Built on our another work EfficientMIL~\cite{she2025_efficientmil}, MMSF replaces quadratic self-attention with the linear-complexity state space model Mamba~\cite{gu2024_mamba}. As illustrated in Figure~\ref{fig:model_pipeline}, MMSF integrates multimodal data through parallel branches: (1) patch-level features extraction using a pathological foundation model with graph construction for spatial context, fused early via a feature fusion module; and (2) instance-level clinical embedding fused late with aggregated histopathological features. This design enables comprehensive multimodal representation learning that combines spatial and clinical insights.

The key contributions of this work are listed as follows:

\begin{itemize}
    \item We propose the first linear-complexity multitask framework for WSI analysis, containing both classification and survival analysis tasks.
    \item We design two plug-and-play modules for extracting patch-level graph features and clinical data features, enhancing multimodal representation learning.
    \item We introduce a hierarchical fusion mechanism: early fusion of graph and pathological features, and late fusion with clinical embeddings.
    \item We achieve state-of-the-art performance across multiple public datasets, validating the effectiveness and generalizability of MMSF.
\end{itemize}

\section{Related Work}
\label{sec:related}

\subsection{WSI Features Extraction}

The evolution of WSI features extraction methods in computational pathology has been marked by a transition from ImageNet-pretrained models~\cite{ilse2018_abmil,li2021_dsmil} to specialized foundation models trained on large-scale histopathology datasets~\cite{chen2024_uni,xu2024_prov_gigapath,zimmermann2024_virchow2,ma2025_gpfm}. Early approaches relied on transfer learning from natural images, but the domain gap between natural and histopathological images limited their effectiveness~\cite{lu2021_clam}. Recent foundation models such as UNI and UNI2~\cite{chen2024_uni} have demonstrated superior performance by training on massive histopathology datasets. These models leverage self-supervised learning on hundreds of millions of patches, enabling robust feature extraction across diverse tissue types and diagnostic tasks.

\subsection{Multiple Instance Learning in WSI Analysis}

Multiple Instance Learning (MIL) has become the standard paradigm for WSI analysis due to the inherent bag-of-patches structure of histopathological images~\cite{ilse2018_abmil}. Traditional attention-based MIL methods, such as ABMIL~\cite{ilse2018_abmil}, learn to focus on the most informative patches within each WSI. Transformer-based approaches like TransMIL~\cite{shao2021_transmil} and Snuffy~\cite{jafarinia2024_snuffy} have achieved significant success by modeling long-range dependencies between patches using self-attention mechanisms. Clustering-based methods such as CLAM~\cite{lu2021_clam} introduced clustering-constrained attention mechanisms for improved patch selection and interpretability. Other prototype-based approaches like PAMIL~\cite{liu2024_pamil} have shown promise in improving interpretability through prototype learning. Recent efficient methods like EfficientMIL~\cite{she2025_efficientmil} proposed linear-complexity alternatives to traditional attention mechanisms. However, these methods still face computational challenges when processing large numbers of patches, and they often ignore the spatial relationships between tissue regions.

\subsection{Graph-based WSI Analysis Methods}

Graph neural networks have shown promise in capturing spatial relationships in medical images~\cite{jaume2021_histocartography}. In computational pathology, graph-based approaches have been used to model tissue architecture and spatial dependencies between patches~\cite{yener_cell_graphs}. HistoCartography~\cite{jaume2021_histocartography} provides a comprehensive toolkit for graph analytics in digital pathology, enabling the construction of cell-graphs and tissue-graphs from histopathology images. Recent work has explored the integration of graph neural networks with multi-omics data for improved prognostic prediction~\cite{kesimoglu2023_supreme,wang2021_mogonet}. Spatial relationship modeling has been particularly important for understanding tumor microenvironment interactions~\cite{shao2024_tumor_micro_environment_interactions, vanea2024_happy}. Graph attention mechanisms have been applied to multi-omics integration tasks~\cite{tanvir2024_mogat}, demonstrating the potential for capturing complex biological relationships.

\subsection{Survival Analysis}

\begin{figure}[!h]
    \centering
    \includegraphics[width=\textwidth]{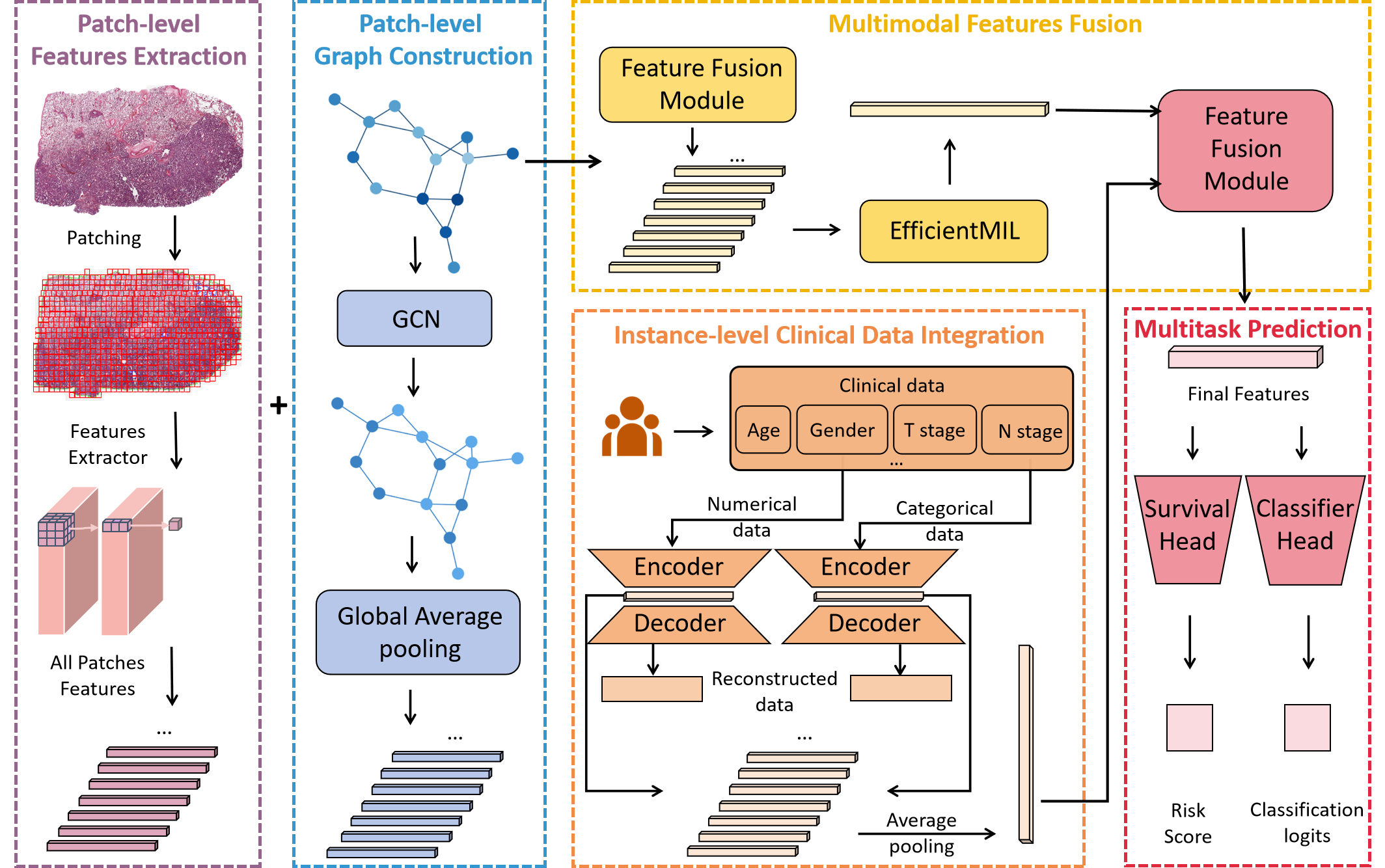}
    \caption{Overview of our proposed MMSF framework. (1) \textbf{Patch-level Features Extraction}. WSIs are cropped into patches and processed by foundation model UNI2 to extract 1536-dimensional patch features. (2) \textbf{Patch-level Graph Construction}. Spatial graphs are constructed based on spatial proximity and tissue similarity, processed by GNN, and aggregated via global average pooling to obtain graph embeddings. (3) \textbf{Multimodal Features Fusion}. Patch features and graph embeddings are fused early using our proposed feature fusion module (FFM), then processed by EfficientMIL with adaptive patch selector (APS) to obtain instance-level representations. And then, the instance-level representations are fused late with clinical embeddings using another FFM to obtain final features. (4) \textbf{Instance-level Clinical Data Integration}. Clinical data (age, gender, T stage, N stage, etc.) is processed separately through our proposed clinical data embedding module (CDE) into clinical embeddings. (5) \textbf{Multitask Prediction}. The final features are fed into task-specific prediction heads (survival head or classifier head) for final predictions.}
    \label{fig:model_pipeline}
\end{figure}

Survival analysis is the statistical study of the time until an event of interest occurs~\cite{wang2019_ml_for_survival_analysis}. In computational pathology, it has evolved from traditional Cox proportional hazards models~\cite{katzman2018_deepsurv} to sophisticated deep learning approaches that can handle high-dimensional data~\cite{lee2018_deephit}. Multi-omics survival prediction has gained significant attention, with methods like MCAT~\cite{chen2021_mcat} and pan-cancer integrative analysis~\cite{chen2022_porpoise} demonstrating the benefits of combining histopathology images with genomic data. WSI-based survival prediction has shown promise in various cancer types~\cite{liu2024_advmil}. Multimodal survival analysis has been particularly successful in capturing complex interactions between different data types~\cite{zhou2023_cross_modal_translation_and_alignment}. However, these approaches often require expensive multi-omics data integration, limiting their clinical applicability due to cost and data availability constraints.

\subsection{Multimodal Fusion Methods}

The integration of multiple data modalities has become increasingly important for improving diagnostic and prognostic accuracy in computational pathology~\cite{chen2021_mcat}. Early fusion approaches combine features at the patch-level, while late fusion strategies integrate information at the bag or instance-level~\cite{li2025_accurate_prediction_disease_free_overall_survival}. Attention-based fusion mechanisms have been particularly successful in capturing complex interactions between different modalities~\cite{chen2021_mcat, xu2024_prov_gigapath}. Recent work has explored multi-level fusion strategies that combine early and late fusion approaches~\cite{yang2025_mmsurv}. However, most existing approaches use single-level fusion strategies, limiting their ability to capture both local and global patterns effectively.

\section{Method}
\label{sec:method}

\subsection{Overview}
\label{subsec:overview}

As shown in Figure~\ref{fig:model_pipeline}, MMSF follows a comprehensive pipeline implemented through the Network class architecture: (1) feature extraction using the pathological foundation model UNI2~\cite{chen2024_uni}, (2) patch-level graph construction and early fusion with WSI features, (3) instance-level classification and Mamba-based MIL encoder with adaptive patch selection~\cite{she2025_efficientmil}, and (4) instance-level clinical data integration and multitask prediction.

Given a WSI with $N$ patches $\{x_1, x_2, ..., x_N\}$, we first extract 1536-dimensional features using UNI2: $f_i = \text{UNI2}(x_i) \in \mathbb{R}^{d_{patch}}$ ($d_{patch}=1536$). The Network class processes these features through multiple stages:

\textbf{Patch-level Graph Construction}. We construct a spatial graph $G = (V, E)$ where vertices $V$ represent patches and edges $E$ connect spatially proximate patches. Graph embeddings $\mathbf{g}_i \in \mathbb{R}^{d_{graph}}$ are processed using graph neural network (GNN).

\textbf{Patch-level Features Fusion}. Extracted patch features are fused with graph embeddings using the Features Fusion Module (FFM), producing fused patch features $\mathbf{f}_{fused}^{(i)} \in \mathbb{R}^{d_{patch^{\prime}}}$, where $d_{patch^{\prime}}$ depends on the fusion strategy.

\textbf{Mamba-based MIL Encoding}. The Mamba-based MIL encoder of EfficientMIL~\cite{she2025_efficientmil} performs instance-level classification on fused patch features and leverages adaptive patches selection strategy to produce both features and instance predictions for subsequent bag-level processing.

\textbf{Instance-level Clinical Data Integration}. Bag representation is fused with clinical embeddings using another new FFM, producing final features for task-specific prediction.

\textbf{Multitask Prediction}. Task-specific heads (classification or survival) process the final fused features to produce predictions.

\subsection{Mamba-based MIL Encoding}
\label{subsec:mamba_based_mil_encoding}

MMSF employs the Mamba-based MIL encoder from EfficientMIL~\cite{she2025_efficientmil} to process patch-level features with linear complexity. Given input patch features $\mathbf{F} = \{\mathbf{f}_1, \mathbf{f}_2, ..., \mathbf{f}_N\}$ where $\mathbf{f}_i \in \mathbb{R}^{d_{patch}}$, the encoder uses adaptive patch selector (APS) to select the most informative patches:

\begin{equation}
\mathbf{F}^{\prime}, \mathbf{S}_{patch} = \text{APS}(\mathbf{F}, \lambda)
\end{equation}

where $\lambda$ controls the number of selected patches, which is set to 512. $\mathbf{S}_{patch}$ is the score of all patches. The selected patches are processed through Mamba2 blocks~\cite{dao2024_mamba2} to produce both instance-level predictions and refined features:

\begin{equation}
\mathbf{F}^{\prime\prime} = \text{EfficientMIL}(\mathbf{F}^{\prime})
\end{equation}

The final bag representation is obtained by aggregating the patch-level graph features and instance-level clinical data features for downstream task-specific prediction heads. This approach achieves linear complexity compared to quadratic complexity of attention-based MIL methods.

\subsection{Patch-level Graph Construction}
\label{subsec:patch_graph_construction}

MMSF constructs spatial graphs to capture tissue architecture and spatial relationships between patches. For patches $i$ and $j$ with spatial coordinates $(x_i, y_i)$ and $(x_j, y_j)$, we define spatial proximity as:
\begin{equation}
d_{spatial}(i,j) = \sqrt{(x_i - x_j)^2 + (y_i - y_j)^2}
\end{equation}

Tissue similarity is measured using cosine similarity between feature vectors:
\begin{equation}
s_{tissue}(i,j) = \frac{f_i \cdot f_j}{\|f_i\| \|f_j\|}
\end{equation}

An edge is created between patches $i$ and $j$ if:
\begin{equation}
d_{spatial}(i,j) < \tau_{spatial} \text{ and } s_{tissue}(i,j) > \tau_{tissue}
\end{equation}

where $\tau_{spatial}$ and $\tau_{tissue}$ are predefined thresholds for spatial proximity and tissue similarity, respectively.

\textbf{Graph Construction}. Given a WSI with $N$ patches, we construct a spatial graph $G = (V, E)$ where:
\begin{itemize}
\item Vertices $V = \{v_1, v_2, ..., v_N\}$ represent patches
\item Edges $E$ connect spatially proximate patches based on tissue similarity
\item Each vertex $v_i$ is associated with extracted patch-level features $f_i \in \mathbb{R}^{d_{patch}}$
\end{itemize}

\textbf{Graph Features Extraction}. The constructed graph is processed using a graph neural network (GNN) to extract spatial context. We conduct an ablation study with different graph models (GCN and GAT) and different parameter combinations for the GNN. Detailed results are available in the supplementary materials, which show that GAT achieves better performance than GCN, and the best parameter combination is $d_{graph\_hidden}=512$ and $d_{graph\_out}=256$.

\begin{equation}
\mathbf{g}_i = \text{GNN}(\mathbf{f}_i, \mathcal{N}(i))
\end{equation}

where $\mathcal{N}(i)$ represents the neighborhood of patch $i$ in the spatial graph, which is processed by the GNN to obtain graph embeddings $\mathbf{g}_i \in \mathbb{R}^{d_{graph}}$.

\subsection{Clinical Data Embedding}
\label{subsec:clinical_module_design}

MMSF incorporates clinical data at the instance level through a dedicated clinical data embedding module (CDE), which is shown in Figure~\ref{fig:model_pipeline}, allowing patient-level clinical features to complement the instance-level representations from histopathological analysis.

\textbf{Module Architecture}. The CDE processes heterogeneous clinical data $\mathbf{c} = [\mathbf{c}_1, \mathbf{c}_2, ..., \mathbf{c}_K]$, where $\mathbf{c}_k \in \mathbb{R}^{1 \times d_{clinical}}$ and each $\mathbf{c}_k$ represents a different clinical data type. For continuous numerical clinical data (e.g., age), $d_{clinical}=1$, and for discrete categorical data (e.g., gender, T stage, N stage, etc.), $d_{clinical}$ is equal to the number of categories. The module uses separate encoder-decoder pairs for each clinical data type to handle the heterogeneity of clinical data. We conduct an experiment with different $d_{clinical}$ values in the supplementary material, which shows that $d_{clinical}=512$ is the best choice.

\textbf{Implementation of Encoder}. For each clinical data $c_k$, we apply a two-layer MLP to extract the hidden representation:
\begin{equation}
\mathbf{h}_k = \mathbf{W}_{k,2} \text{ReLU}(\mathbf{W}_{k,1} \mathbf{c}_k + \mathbf{b}_{k,1}) + \mathbf{b}_{k,2}
\end{equation}
where $\mathbf{h}_k \in \mathbb{R}^{d_{hidden}}$ is the hidden representation.

\textbf{Implementation of Decoder}. The decoder reconstructs the original features for self-supervised learning:
\begin{equation}
\hat{\mathbf{c}}_k = \mathbf{W}_{k,4} \text{ReLU}(\mathbf{W}_{k,3} \mathbf{h}_k + \mathbf{b}_{k,3}) + \mathbf{b}_{k,4}
\end{equation}
where $\hat{\mathbf{c}}_k \in \mathbb{R}^{1 \times d_{clinical}}$ is the reconstructed clinical data.

\textbf{Feature Aggregation}. The final clinical embedding is obtained by averaging the hidden representations:
\begin{equation}
\mathbf{c}_{emb} = \frac{1}{K} \sum_{k=1}^{K} \mathbf{h}_k
\end{equation}

\textbf{Reconstruction Loss}: The reconstruction loss encourages the clinical embedding to preserve important information:
\begin{equation}
\label{eq:clinical_loss}
L_{clinical} = \sum_{k=1}^{K} \mathcal{L}(\hat{\mathbf{c}}_k, \mathbf{c}_k)
\end{equation}

where $\mathcal{L}$ is the appropriate loss function:
\begin{equation}
\mathcal{L}(\hat{\mathbf{c}}_k, \mathbf{c}_k) = \begin{cases}
\text{MSE}(\hat{\mathbf{c}}_k, \mathbf{c}_k) & \text{if data is numerical} \\
\text{CrossEntropy}(\hat{\mathbf{c}}_k, \mathbf{c}_k) & \text{if data is categorical}
\end{cases}
\end{equation}

\subsection{Features Fusion Module}
\label{subsec:ffm_design}

We propose a novel Features Fusion Module (FFM) for both patch-level and instance-level integration, supporting multiple fusion strategies optimized for different contexts. The FFM enables flexible and effective combination of heterogeneous features, as shown in Figure~\ref{fig:model_pipeline}. We also conduct an ablation study with different fusion strategies in the supplementary material, which shows that SE-based~\cite{hu2018_senet} fusion achieves better performance than other fusion strategies.

\textbf{Patch-level Early Fusion}. For graph integration, we use early fusion to combine patch features with graph embedding features before bag aggregation. This allows spatial relationships to influence patch-level representations during the learning process.

\textbf{Instance-level Late Fusion}. For clinical integration, we use late fusion to combine bag-level representations with clinical embedding features after MIL processing. This preserves the hierarchical nature of the data while incorporating patient-level information.

\textbf{Implementation of FFM}. The FFM takes two input features $\mathbf{f}_1 \in \mathbb{R}^{d_1}$ and $\mathbf{f}_2 \in \mathbb{R}^{d_2}$ and produces fused features $\mathbf{f}_{fused} \in \mathbb{R}^{d_1 + d_2}$. The implementation of FFM is shown in the following formulas:

\begin{equation}
\mathbf{f}_{fused} = LayerNorm(SE(concat(\mathbf{f}_1, \mathbf{f}_2)))
\end{equation}

First, concatenate $\mathbf{f}_1$ and $\mathbf{f}_2$ together. Then, fuse the concatenated feature vectors using the SE attention mechanism. Finally, apply LayerNorm to the fused features.

\subsection{Multitask Prediction Heads}
\label{subsec:multitask_heads_design}

MMSF employs separate prediction heads for classification and survival analysis tasks, allowing the model to learn task-specific representations while sharing common features. The heads are applied to the final fused features $\mathbf{z}$ after all fusion operations.

\textbf{Classification Head}. For WSI classification tasks, we use just a linear layer:
\begin{equation}
\mathbf{p}_{class} = \mathbf{W}_{class} \mathbf{z} + \mathbf{b}_{class}
\end{equation}
where $\mathbf{z}$ is the final fused representation and $\mathbf{W}_{class}, \mathbf{b}_{class}$ are learnable parameters.

\textbf{Survival Head}. For survival analysis, we use a linear layer followed by sigmoid activation to predict risk scores:
\begin{equation}
r = \sigma(\mathbf{W}_{surv} \mathbf{z} + \mathbf{b}_{surv})
\end{equation}
where $\sigma$ is the sigmoid function and $r \in [0,1]$ represents the risk score.

\textbf{Head Selection}. The appropriate head is selected based on the task type:
\begin{equation}
\mathbf{y} = \begin{cases}
\text{ClassificationHead}(\mathbf{z}) & \text{if task is classification} \\
\text{SurvivalHead}(\mathbf{z}) & \text{if task is survival analysis}
\end{cases}
\end{equation}

\subsection{Details of Loss Functions}
\label{subsec:details_loss_function}

MMSF employs a comprehensive loss function design that combines multiple objectives to enable effective multitask learning. The total loss integrates classification, survival, and clinical reconstruction losses.

\textbf{Classification Loss.} For classification tasks, we first obtain the bag-level representation through global average pooling of the instance-level features. Given $\lambda$ selected instances with refined features $\mathbf{H}_{res} = \{\mathbf{h}_1, \mathbf{h}_2, ..., \mathbf{h}_\lambda\}$ where $\mathbf{h}_i \in \mathbb{R}^{1 \times d}$, the bag representation is computed as:
\begin{equation}
\mathbf{z}_{bag} = \frac{1}{\lambda} \sum_{i=1}^{\lambda} \mathbf{H}_{res}[i]
\end{equation}
where $\mathbf{z}_{bag} \in \mathbb{R}^{1 \times d}$ is the bag-level representation.

The final classification logits are computed as:
\begin{equation}
\hat{\mathbf{y}} = \mathbf{W}_{cls} \mathbf{z}_{bag} + \mathbf{b}_{cls}
\end{equation}
where $\mathbf{W}_{cls} \in \mathbb{R}^{C \times d}$ and $\mathbf{b}_{cls} \in \mathbb{R}^{C}$ are the classification layer weight matrix and bias vector, respectively, and $C$ is the number of classes (In our experiments, $C=2$ for binary classification).

Our training objective for classification uses binary cross-entropy with logits (BCEWithLogits) on two terms: the bag-level prediction and the max-pooled instance prediction, combined with equal weights. The classification loss is defined as:
\begin{equation}
L_{cls} = \frac{1}{2} \text{BCELogits}(\hat{\mathbf{y}}_{bag}, \mathbf{y}) + \frac{1}{2} \text{BCELogits}(\max_i \hat{\mathbf{y}}_{inst,i}, \mathbf{y})
\end{equation}
where $\hat{\mathbf{y}}_{bag}$ is the bag-level prediction, $\hat{\mathbf{y}}_{inst,i}$ is the instance-level prediction for the $i$-th instance, $\mathbf{y}$ is the ground truth label, and $\max_i \hat{\mathbf{y}}_{inst,i}$ denotes taking the maximum logit across all instances within the bag.

\textbf{Survival Analysis Loss.} For survival analysis tasks, we employ the Cox proportional hazards model implemented through the negative partial log-likelihood function~\cite{cox1972_cox_regression}. The risk score for patient $i$ is computed as:
\begin{equation}
r_i = \sigma(\mathbf{W}_{surv} \mathbf{z}_i + \mathbf{b}_{surv})
\end{equation}
where $\mathbf{z}_i$ is the patient's feature representation and $\sigma$ is the sigmoid function producing risk scores in $[0,1]$.

The survival loss is computed as:
\begin{equation}
L_{surv} = -\frac{1}{N} \sum_{i=1}^{N} \delta_i \left[ r_i - \log \sum_{j \in R_i} \exp(r_j) \right] + \lambda_{reg} \|\mathbf{r}\|_2
\end{equation}
where $\delta_i$ is the event indicator (1 for events, 0 for censored), $R_i$ is the risk set for patient $i$ (patients with survival times $\geq t_i$), and $\lambda_{reg}$ is a regularization term, which is set to $1 \times 10^{-4}$.

\textbf{Total Loss.} The total loss function combines the task-specific losses with the clinical reconstruction loss and L2 regularization:
\begin{equation}
L_{total} = L_{task} + \lambda_{L2} \|\boldsymbol{\theta}\|_2^2
\end{equation}
where $L_{task}$ is either $L_{cls}$ for classification tasks or $L_{surv}$ for survival analysis tasks, $\lambda_{L2}$ is the L2 regularization coefficient set to $1 \times 10^{-4}$, and $\boldsymbol{\theta}$ represents all model parameters.

\section{Results}
\label{sec:results}

\subsection{Experimental Settings}

\textbf{Datasets.} We evaluate MMSF on both classification and survival analysis tasks across multiple datasets. For the classification task, we use CAMELYON16 and TCGA-NSCLC datasets~\cite{bejnordi2017_camelyon16,cooper2018_tcga_pancancer}. For survival analysis tasks, we use five TCGA datasets~\cite{cooper2018_tcga_pancancer}: TCGA-BLCA, TCGA-COAD, TCGA-LUAD, TCGA-STAD, and TCGA-KIRC. More details about these datasets are provided in the supplementary material.

\textbf{Evaluation Metrics.} For classification tasks on CAMELYON16 and TCGA-NSCLC, we use accuracy (ACC) and area under the ROC curve (AUC) to evaluate binary classification performance. For survival analysis tasks on all five TCGA datasets, we use the concordance index (C-index) to assess survival risk ranking. Computational efficiency is measured by FLOPs (in megaflops (M)) and model size (in megabytes (MB)), as shown in Tables~\ref{tab:classification_results} and~\ref{tab:survival_results}.

\begin{figure}[!ht]
    \centering
    \includegraphics[width=0.9\textwidth]{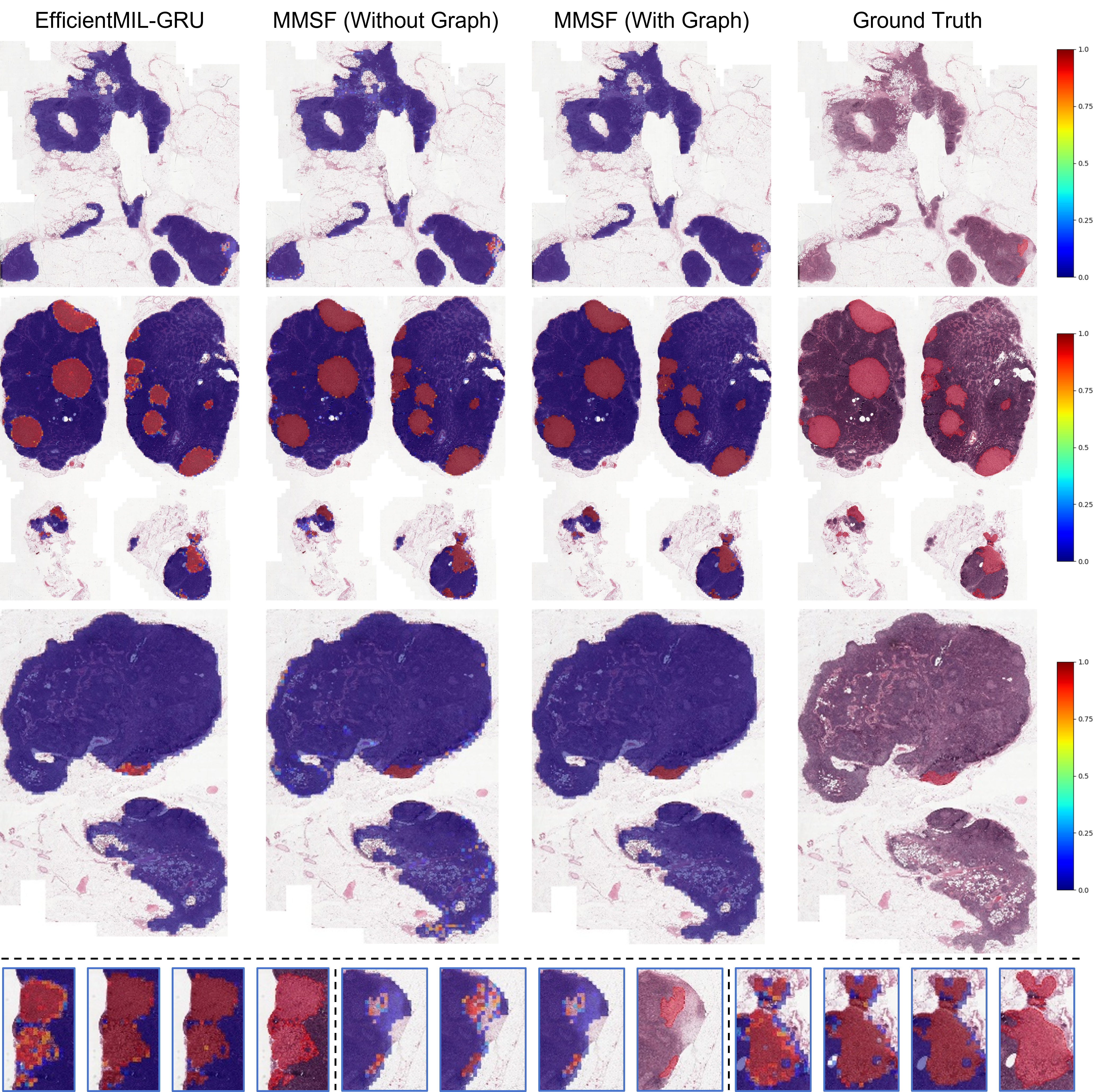}
    \caption{Visualization of patch scores from the adaptive patch selector (APS) on CAMELYON16 dataset for classification task. The heatmaps overlay patch scores $S_{patch}$ on WSI, where red regions (score $\approx$ 1.0) indicate highly informative patches selected by APS, and blue regions (score $\approx$ 0.0) represent less informative patches.}
    \label{fig:patch_score_visualization}
\end{figure}

\textbf{Implementation Details.} All experiments are conducted with 2$\times$NVIDIA 3090 GPUs using PyTorch 2.11+cu118 on Ubuntu 20.04.6 LTS. We use PyTorch Geometric 2.6.1 for graph neural network operations and torchmetrics 1.7.3 for evaluation metrics. For reproducibility, we set the random seed to 42. For both classification and survival analysis tasks, we train for 50 epochs using the Adam optimizer with learning rate $2 \times 10^{-4}$ and weight decay $1 \times 10^{-5}$. The learning rate scheduler uses warmup for 5 epochs followed by step decay with $\gamma=0.6$ and minimum learning rate $1 \times 10^{-8}$. The batch size is set to 1 WSI per GPU due to memory constraints. We employ an 80\%/20\% train/validation split for all datasets. Early stopping is applied after 6 epochs without validation improvement. For survival analysis tasks, we also use Cox proportional hazards loss and C-index evaluation. More details about model hyperparameters are provided in the supplementary material.

\begin{table}[h]
    \centering
    \caption{Performance comparison of different methods for WSI classification tasks. Values are reported as the result on 20\% validation set. Best and second-best results are highlighted in bold and underlined, respectively.}
    \label{tab:classification_results}
    \setlength{\tabcolsep}{2pt}
    \begin{tabular}{lcccc}
    \toprule
    \textbf{Methods} 
        & \multicolumn{2}{c}{\textbf{TCGA-NSCLC}} 
        & \multicolumn{2}{c}{\textbf{CAMELYON16}} \\
    \cmidrule(lr){2-3} \cmidrule(lr){4-5}
        & {\textbf{ACC}} & {\textbf{AUC}} & {\textbf{ACC}} & {\textbf{AUC}} \\
    \midrule
    ABMIL~\cite{ilse2018_abmil}                 & 0.812  & 0.853  & 0.835  & 0.873 \\
    DGMIL~\cite{qu2022_dgmil}                 & 0.601  & 0.632  & 0.620  & 0.659 \\
    TransMIL~\cite{shao2021_transmil}              & 0.899  & 0.911  & 0.906  & 0.926 \\
    DTFD-MIL~\cite{zhang2022_dtfd_mil}       & 0.901  & 0.912  & 0.912  & 0.925 \\
    Snuffy~\cite{jafarinia2024_snuffy}                & 0.934  & 0.961  & \underline{0.947}  & 0.956 \\
    CAMIL~\cite{mao2025_camil}                 & \underline{0.953}  & \textbf{0.985}  & 0.915  & \underline{0.975} \\
    EfficientMIL~\cite{she2025_efficientmil} & 0.933  & 0.976  & 0.823  & 0.933 \\
    \hline
    \textbf{MMSF (Ours)}          & \textbf{0.957} & \underline{0.981} & \textbf{0.988} & \textbf{0.994} \\
    \bottomrule
\end{tabular}
\end{table}

\subsection{Classification Results}

We compare MMSF against several baseline methods on both tasks. Table~\ref{tab:classification_results} presents the performance comparison for the classification task on CAMELYON16 and TCGA-NSCLC datasets. Compared to existing MIL methods, MMSF achieves significant improvements with accuracy gains of 2.1--6.6\% and AUC improvements of 2.2--6.9\% across both classification datasets.

To provide deeper insights into MMSF's performance, we visualize $S_{patch} \in \mathbb{R}^{N \times d_{patch}}$ (where $N$ is the number of patches cropped from the WSI), which is introduced in Section~\ref{subsec:mamba_based_mil_encoding}, for the classification task on the CAMELYON16 dataset. The results are shown in Figure~\ref{fig:patch_score_visualization}. They demonstrate that MMSF effectively identifies and focuses on diagnostically relevant tissue regions, with high-scoring patches concentrated in tumor areas and critical morphological structures.

\subsection{Survival Analysis Results}

The performance comparison for the survival analysis task on five TCGA datasets is shown in Table~\ref{tab:survival_results}. MMSF achieves the best performance on four out of five datasets (TCGA-BLCA, TCGA-COAD, TCGA-LUAD, and TCGA-STAD), with an overall C-index of 0.6911, representing improvements of 8.9\% and 5.5\% compared to the best unimodal (MLP, 0.6345) and multimodal (HSFSurv, 0.6552) baseline methods, respectively. The results demonstrate that MMSF achieves competitive performance across diverse cancer types.

\begin{table}[!ht]
    \centering
    \setlength{\tabcolsep}{2pt}
    \tiny
    \caption{Performance comparison of different methods for survival analysis task on five TCGA datasets. Values denote the C-index on 20\% validation set. The last \textbf{Overall} column is the average C-index on all five datasets. Best and second-best results per column are highlighted in bold and underlined, respectively.}
    \label{tab:survival_results}
    \adjustbox{width=\textwidth,center}{\begin{tabular}{lcccccc}
    \toprule
    \textbf{Methods} & \textbf{TCGA-BLCA} & \textbf{TCGA-COAD} & \textbf{TCGA-LUAD} & \textbf{TCGA-STAD} & \textbf{TCGA-KIRC} & \textbf{Overall} \\
    \midrule
    \multicolumn{7}{l}{\textit{Unimodal Methods}} \\
    SNN~\cite{yue2018_snn} & $0.6523$ & $0.6189$ & $0.5934$ & $0.5718$ & $0.7106$ & $0.6294$ \\
    MLP & $0.6845$ & $0.5821$ & $0.5998$ & $0.5972$ & $0.7087$ & $0.6345$ \\
    ABMIL~\cite{ilse2018_abmil} & $0.4918$ & $0.4967$ & $0.5959$ & $0.5672$ & $0.5669$ & $0.5437$ \\
    TransMIL~\cite{shao2021_transmil} & $0.5065$ & $0.5018$ & $0.5781$ & $0.5637$ & $0.5752$ & $0.5451$ \\
    CLAM-SB~\cite{lu2021_clam} & $0.5376$ & $0.5032$ & $0.5873$ & $0.5641$ & $0.5964$ & $0.5577$ \\
    \midrule
    \multicolumn{7}{l}{\textit{Multimodal Methods}} \\
    MCAT~\cite{chen2021_mcat} & $\underline{0.6998}$ & $0.5837$ & $\underline{0.6287}$ & $0.6261$ & $0.7019$ & $0.6480$ \\
    CMTA~\cite{zhou2023_cmta} & $0.6538$ & $0.5859$ & $0.5663$ & $0.6252$ & $0.7067$ & $0.6276$ \\
    SurvPath~\cite{jaume2024_survpath} & $0.6856$ & $0.6371$ & $0.5925$ & $0.6248$ & $0.6991$ & $0.6478$ \\
    CCL~\cite{zhou2025_ccl} & $0.6795$ & $\underline{0.6463}$ & $0.5803$ & $0.5937$ & $0.7076$ & $0.6415$ \\
    Pathomics~\cite{ding2023_pathomics} & $0.6049$ & $0.5367$ & $0.5421$ & $\underline{0.6295}$ & $0.6934$ & $0.6013$ \\
    HSFSurv~\cite{fu2025_hsfsurv} & $0.6832$ & $0.6456$ & $0.6165$ & $0.6018$ & $\textbf{0.7287}$ & $\underline{0.6552}$ \\
    \textbf{MMSF (Ours)} & $\textbf{0.7024}$ & $\textbf{0.6934}$ & $\textbf{0.6625}$ & $\textbf{0.6721}$ & $\underline{0.7251}$ & $\textbf{0.6911}$ \\
    \bottomrule
\end{tabular}}
\end{table}

\begin{figure}[!ht]
    \centering
    \includegraphics[width=0.9\textwidth]{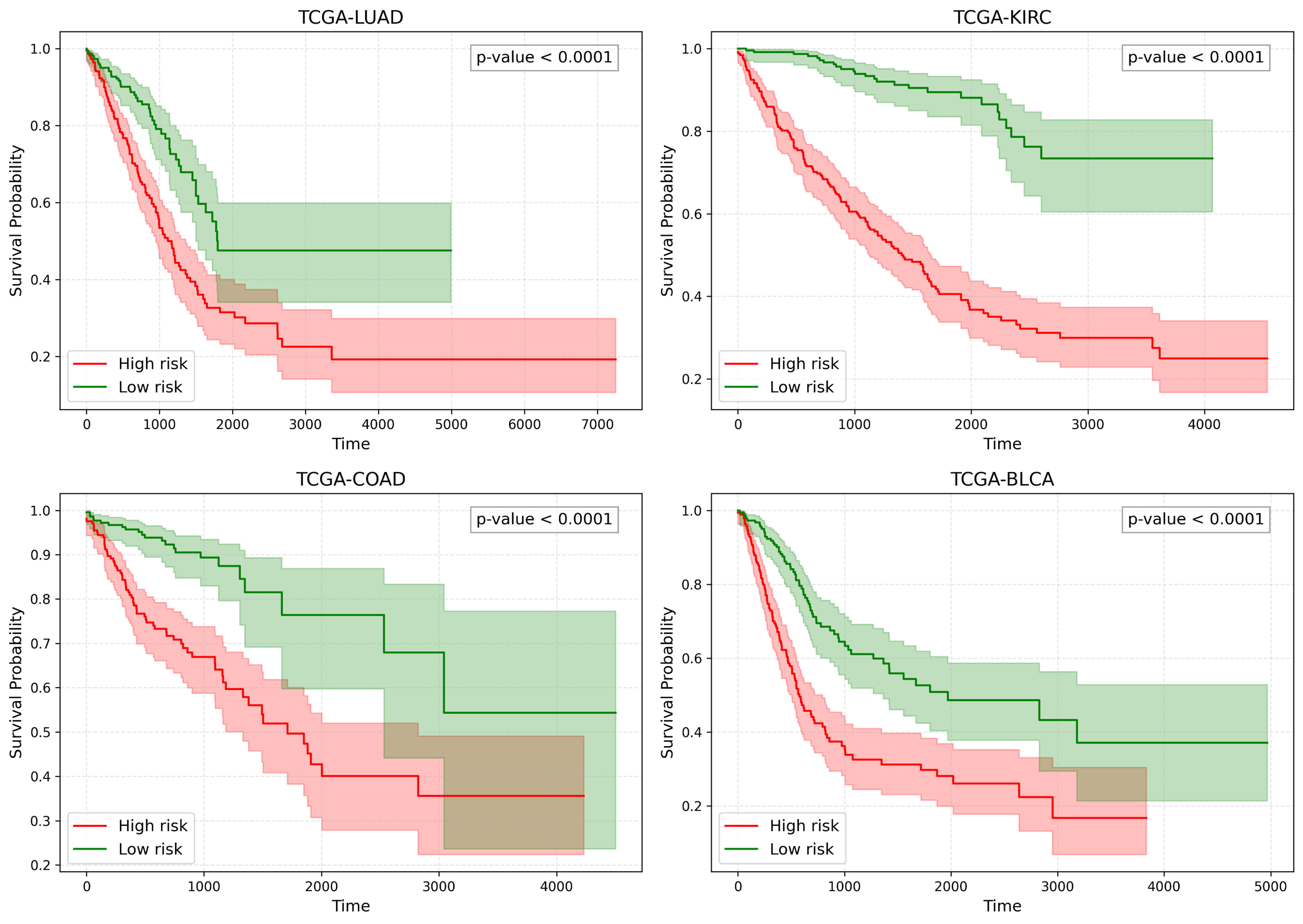}
    \caption{Kaplan-Meier survival curves for high-risk and low-risk patient groups stratified by MMSF predictions across four TCGA datasets: TCGA-LUAD (top-left), TCGA-KIRC (top-right), TCGA-COAD (bottom-left), and TCGA-BLCA (bottom-right). Patients are divided into high-risk (red) and low-risk (green) groups based on predicted survival risk scores. Shaded regions represent 95\% confidence intervals.}
    \label{fig:kaplan_meier_visualization}
\end{figure}

Survival curve stratification further validates MMSF's effectiveness, as shown in Figure~\ref{fig:kaplan_meier_visualization}. MMSF successfully separates patients into high-risk and low-risk groups with statistically significant differences in survival outcomes across four TCGA datasets (log-rank test, $p < 0.0001$). This highlights the model's ability to integrate multimodal data and shared representations for robust multitask performance.

\subsection{Ablation Studies}
\label{subsec:ablation}

\textbf{Effect of Graph Construction.} Table~\ref{tab:classification_ablation} quantifies the benefit of introducing patch-level spatial graphs. Adding the graph branch improves ACC and AUC on both TCGA-NSCLC and CAMELYON16 datasets. These improvements are achieved with only a moderate increase in model size, indicating that the spatial graph effectively injects contextual tissue relationships with limited additional parameters. Table~\ref{tab:survival_ablation} also shows the benefit of introducing patch-level spatial graphs for survival analysis by comparing the backbone model with models incorporating patch-level spatial graph features (+ Graph).

\begin{table}[!ht]
    \centering
    \caption{Performance comparison of the backbone model (Baseline) with models incorporating patch-level spatial graph features (+ Graph). Values denote the accuracy and AUC on 20\% validation set.}
    \label{tab:classification_ablation}
    \setlength{\tabcolsep}{2pt}
    \begin{tabular}{lcccccc}
    \toprule
    \textbf{Methods} 
        & \multicolumn{2}{c}{\textbf{TCGA-NSCLC}} 
        & \multicolumn{2}{c}{\textbf{CAMELYON16}} 
        & \textbf{Model Size} \\
    \cmidrule(lr){2-3} \cmidrule(lr){4-5}
        & {\textbf{ACC}} & {\textbf{AUC}} & {\textbf{ACC}} & {\textbf{AUC}} & {\textbf{(MB)}} \\
    \midrule
    Baseline & 0.9372 & 0.9763 & 0.9575 & 0.9786 & 436 \\
    + Graph & 0.9476 & 0.9905 & 0.9975 & 0.9988 & 595 \\
    \bottomrule
\end{tabular}
\end{table}

\textbf{Effect of Clinical Data Embedding.} Table~\ref{tab:survival_ablation} isolates the effect of graph and clinical cues for survival prediction. Incorporating the spatial graph alone boosts the C-index on all five datasets relative to the backbone. Additional ablation studies of hyperparameters are provided in the supplementary material.
    
\begin{table}[!ht]
    \centering
    \small
    \caption{Performance comparison of the backbone model (Baseline) with models incorporating patch-level spatial graph features (+ Graph), instance-level clinical data embedding via CDE module (+ Clinical), and both components combined (+ Both). Values denote the C-index on 20\% validation set.}
    \label{tab:survival_ablation}
    \begin{tabular}{lccc}
    \toprule
    \textbf{Methods} & \textbf{TCGA-COAD} & \textbf{TCGA-LUAD} & \textbf{TCGA-STAD} \\
    \midrule
    Baseline & 0.6108 & 0.5637 & 0.5870 \\
    + Graph & 0.6748 & 0.6481 & 0.6314  \\
    + Clinical & 0.6854 & 0.6518 & 0.6693 \\
    + Both & 0.6934 & 0.6625 & 0.6721 \\
    \bottomrule
\end{tabular}
\end{table}
    
\textbf{Effect of Features Fusion Module.} Table~\ref{tab:fusion_ablation} compares three fusion strategies inside the proposed FFM. The SE-based fusion strategy has the best performance on all three datasets over linear fusion or no fusion.

\begin{table}[!ht]
    \centering
    \small
    \caption{Performance comparison of different fusion strategies for multimodal features integration. The table evaluates None (no fusion), Linear fusion, and SE-based fusion strategies. Values denote the C-index on 20\% validation set.}
    \label{tab:fusion_ablation}
    \begin{tabular}{lccc}
    \toprule
    \textbf{Methods} & \textbf{TCGA-COAD} & \textbf{TCGA-LUAD} & \textbf{TCGA-STAD} \\
    \midrule
    None & 0.6108 & 0.6236 & 0.5932 \\
    Linear & 0.6575 & 0.6481 & 0.6314  \\
    SE & 0.6934 & 0.6625 & 0.6721 \\
    \bottomrule
\end{tabular}
\end{table}

These ablation results confirm that each component in MMSF contributes meaningfully to the overall performance, with the combination of spatial graph constructor, CDE module, and SE-based FFM achieving the best results across both classification and survival analysis tasks. More details of the ablation studies are provided in the supplementary material.

\section{Discussion}
\label{sec:discussion}

In this paper, we present MMSF, a novel Multitask and Multimodal Supervised Framework for WSI classification and survival analysis tasks that integrates multimodal WSI patch-level graph structure information and instance-level clinical data features into extracted WSI features. We also propose a novel features fusion module and employ linear-complexity MIL algorithm to effectively reduce computation compared to conventional transformer-based MIL methods.

\subsection{Comparison with Existing Methods}

MMSF demonstrates significant improvements over existing methods across multiple evaluation metrics. Compared to unimodal approaches such as TransMIL~\cite{shao2021_transmil} and CLAM~\cite{lu2021_clam}, MMSF achieves superior performance by incorporating both spatial graph information and clinical data. The integration of multimodal features enables the model to capture complementary information that is not available in image-only approaches.

Compared to existing multimodal methods such as MCAT~\cite{chen2021_mcat} and HSFSurv~\cite{fu2025_hsfsurv}, MMSF's hierarchical fusion strategy (early fusion for graph features and late fusion for clinical embeddings) provides a more effective way to combine heterogeneous modalities. The linear-complexity Mamba-based backbone also offers computational advantages over transformer-based approaches, making MMSF more suitable for clinical deployment.

\subsection{Clinical Significance}

The computational efficiency and clinical practicality make MMSF a promising tool for advancing multitask analysis in computational pathology, playing a crucial role in clinical applications. The ability to simultaneously perform classification and survival analysis using routinely available WSI and clinical data (without requiring expensive multi-omics data) addresses a key barrier to clinical translation. The model's interpretability through patch-level attention visualization also provides valuable insights for pathologists.

\subsection{Limitations and Future Work}

Several limitations should be acknowledged. First, the graph construction step, while using spatial proximity constraints, may still have computational overhead for very large WSIs. Future work could explore more efficient graph construction algorithms or approximate nearest neighbor methods to further optimize this step. Second, the current implementation assumes complete clinical data availability. Handling missing clinical features robustly remains an important direction for future research.

Third, while we evaluated MMSF on multiple cancer types, the generalizability to other disease domains and imaging modalities warrants further investigation. Additionally, the current framework processes each WSI independently. Extending the model to leverage patient-level longitudinal data or multiple slides per patient could further improve prognostic accuracy.

Finally, while we demonstrated improvements over existing methods, a more comprehensive comparison with recent foundation model-based approaches and larger-scale validation on independent test sets would strengthen the clinical validity of our findings.

\subsection{Conclusions}

We have introduced MMSF, a multitask and multimodal supervised framework for WSI classification and survival analysis. The key innovations include: (1) a linear-complexity Mamba-based MIL backbone that efficiently processes large numbers of patches, (2) a patch-level graph construction module that captures spatial tissue relationships, (3) a clinical data embedding module that integrates heterogeneous patient attributes, and (4) a hierarchical fusion mechanism that effectively combines multimodal features at different levels.

Experimental results on seven public datasets demonstrate that MMSF achieves state-of-the-art performance on both classification and survival analysis tasks, with significant improvements over existing unimodal and multimodal methods. The ablation studies confirm that each component contributes meaningfully to the overall performance. These results position MMSF as a scalable and reliable solution for multimodal computational pathology, with potential for clinical translation.

\bibliographystyle{elsarticle-num}
\bibliography{main}

\appendix

\section{Supplementary Material}
\label{sec:appendix}

\subsection{Datasets for Classification and Survival Tasks}
\label{subsec:datasets_details}

This study uses publicly available datasets without direct human participant involvement. All data were previously collected under appropriate IRB approvals and informed consent as described in the original publications; we only use de-identified data and follow the licenses and usage terms reported by the dataset providers.

\textbf{Classification task datasets.} CAMELYON16~\cite{bejnordi2017_camelyon16} consists of 399 WSIs from 233 breast cancer patients with normal tissue and lymph node metastases. Following common practice, we extract fixed-scale tiles from tissue regions and compute patch-level features using the pathological foundation model UNI2~\cite{chen2024_uni}; patches outside tissue masks are discarded. TCGA-NSCLC (LUAD vs. LUSC) is curated from TCGA~\cite{cooper2018_tcga_pancancer} and contains WSIs from patients with lung adenocarcinoma (LUAD) and lung squamous cell carcinoma (LUSC).

\textbf{Survival analysis task datasets.} We evaluate on five TCGA cohorts~\cite{cooper2018_tcga_pancancer}: BLCA, COAD, LUAD, STAD, and KIRC. For each patient, we use the survival time and event indicator and aggregate slide-level information to the patient level. Clinical variables (if used) follow the dataset CSVs; numeric columns are z-score normalized and categorical columns are one-hot encoded. Patients with missing essential survival fields are excluded from the corresponding fold.

\subsection{Model and Training Hyperparameters}
\label{subsec:hparam_details}

The complete hyperparameter settings for classification and survival tasks are shown in Table~\ref{tab:hparam_summary}. All settings are aligned with our released code.

\begin{table}[!ht]
    \centering
    \caption{Complete hyperparameter settings for classification and survival tasks.}
    \label{tab:hparam_summary}
    \setlength{\tabcolsep}{4pt}
    \scriptsize
    \begin{tabular}{lcc}
    \toprule
    \textbf{Hyperparameter} & \textbf{Classification} & \textbf{Survival} \\
    \midrule
    \multicolumn{3}{l}{\textit{Common Settings}} \\
    \midrule
    Patch feature size & 1536 & 1536 \\
    Dropout & 0.1 & 0.1 \\
    Epochs & 50 & 50 \\
    Optimizer & Adam & Adam \\
    Learning rate & $1\times10^{-4}$ & $1\times10^{-4}$ \\
    Weight decay & $1\times10^{-5}$ & $1\times10^{-5}$ \\
    Scheduler & Warmup(5)+Step & Warmup(5)+Step \\
    Step size & 2 & 2 \\
    Scheduler $\gamma$ & 0.6 & 0.6 \\
    Minimum learning rate & $1\times10^{-8}$ & $1\times10^{-8}$ \\
    Batch size (WSI/GPU) & 1 & 1 \\
    Train/Validation split & 80\%/20\% & 80\%/20\% \\
    Early stopping patience & 10 & 10 \\
    Random seed & 42 & 42 \\
    \midrule
    \multicolumn{3}{l}{\textit{Mamba MIL Encoder}} \\
    \midrule
    Depth & 8 & 8 \\
    State dimension & 16 & 16 \\
    Convolution width & 4 & 4 \\
    Expansion factor & 2 & 2 \\
    Big lambda & 512 & 512 \\
    \midrule
    \multicolumn{3}{l}{\textit{Graph Encoder (Optional)}} \\
    \midrule
    Graph model & \texttt{GAT} & \texttt{GAT} \\
    Graph hidden dim & 256 & 256 \\
    Graph output dim & 256 & 256 \\
    Graph dropout & 0.1 & 0.1 \\
    \midrule
    \multicolumn{3}{l}{\textit{Clinical Encoder (Optional, Survival Analysis Task Only)}} \\
    \midrule
    Clinical hidden dim & N/A & 512 \\
    Reconstruction loss weight & N/A & 0.1 \\
    Numeric normalization & N/A & z-score \\
    Categorical encoding & N/A & one-hot \\
    \midrule
    \multicolumn{3}{l}{\textit{Regularization}} \\
    \midrule
    L2 regularization & enabled & enabled \\
    L2 $\lambda$ (initial) & $1\times10^{-6}$ & $1\times10^{-6}$ \\
    L2 schedule & decreasing & decreasing \\
    L2 $\lambda$ (max) & $\leq 1\times10^{-4}$ & $\leq 1\times10^{-4}$ \\
    L2 penalty target & classifier layers & classifier layers \\
    \bottomrule
\end{tabular}
\end{table}

\subsection{Additional Ablation Studies}
\label{subsec:additional_results}

\subsubsection{Extended Ablation Studies}
\label{subsubsec:extended_ablation}

\textbf{Effect of the hyperparameters in Graph Construction.} Table~\ref{tab:ablation_graph_params} reports the C-index performance when varying the hidden dimension ($d_{hidden}$) and output dimension ($d_{out}$) of the spatial graph encoder (we use GAT as the graph model). The evaluation is conducted on three TCGA cohorts (COAD, LUAD, STAD) using the 20\% validation split. The configuration with $d_{hidden}=256$ and $d_{out}=256$ yields the best trade-off across all datasets.

\begin{table}[!ht]
    \centering
    \scriptsize
    \caption{Impact of graph encoder dimensions on survival analysis. Each entry reports the C-index on the 20\% validation set when using the specified $d_{hidden}$ and $d_{out}$ dimensions.}
    \label{tab:ablation_graph_params}
    \begin{tabular}{ccccc}
    \toprule
    $d_{hidden}$ & $d_{out}$ & \textbf{TCGA-COAD} & \textbf{TCGA-LUAD} & \textbf{TCGA-STAD} \\
    \midrule
    128 & 128 & 0.6271 & 0.5726 & 0.6174 \\
    256 & 128 & 0.6526 & 0.6162 & 0.6221 \\
    256 & 256 & \textbf{0.6748} & \textbf{0.6481} & \textbf{0.6314} \\
    512 & 256 & 0.6014 & 0.6027 & 0.6127 \\
    512 & 512 & 0.6421 & 0.5664 & 0.5969 \\
    \bottomrule
\end{tabular}
\end{table}

\textbf{Effect of the hyperparameters in Clinical Data Embedding.} Table~\ref{tab:ablation_clinical_params} reports the C-index performance when varying the hidden dimension ($d_{hidden}$) of the clinical embedding module. The evaluation is conducted on three TCGA cohorts (BLCA, COAD, LUAD) using the 20\% validation split. The configuration with $d_{hidden}=512$ yields the best performance on all three datasets.

\begin{table}[!ht]
    \centering
    \small
    \caption{Clinical hidden dimension sweep for the CDE module. Values denote the C-index on the validation split for each dataset.}
    \label{tab:ablation_clinical_params}
    \begin{tabular}{lccc}
    \toprule
    $d_{hidden}$ & \textbf{TCGA-COAD} & \textbf{TCGA-LUAD} & \textbf{TCGA-STAD} \\
    \midrule
    128 & 0.6350 & 0.6015 & 0.6258 \\
    256 & 0.6482 & 0.6114 & 0.6492 \\
    512 & \textbf{0.6854} & \textbf{0.6518} & \textbf{0.6693} \\
    1024 & 0.6246 & 0.6117 & 0.6688 \\
    \bottomrule
\end{tabular}
\end{table}

\section*{Conflict of Interest}
The authors declare that they have no known competing financial interests or personal relationships that could have appeared to influence the work reported in this paper.

\section*{Ethical Approval}
This study uses publicly available datasets without direct human participant involvement. All data were previously collected under appropriate IRB approvals and informed consent as described in the original publications; we only use de-identified data and follow the licenses and usage terms reported by the dataset providers.

\end{document}